\documentclass[10pt,letterpaper]{article}

\usepackage{amsmath}
\usepackage{ccn}
\usepackage{pslatex}
\usepackage{apacite}
\usepackage{verbatim}
\usepackage{graphicx}
\usepackage[colorinlistoftodos,prependcaption,textsize=tiny]{todonotes}
\usepackage{booktabs}

\usepackage{array}
\newcolumntype{C}[1]{>{\centering\let\newline\\\arraybackslash\hspace{0pt}}m{#1}}
\newcolumntype{L}[1]{>{\let\newline\\\arraybackslash\hspace{0pt}}m{#1}}

\title{A First Step in Combining Cognitive Event Features and Natural Language Representations to Predict Emotions} 

\author{{\large \bf Andres Campero (campero@mit.edu)} \\
  {\large \bf Bjarke Felbo (felbo@mit.edu)}\\ 
  {\large \bf Joshua B. Tenenbaum (jbt@mit.edu)} \\
  {\large \bf Rebecca Saxe (saxe@mit.edu)} \\
  Massachusetts Institute of Technology, Cambridge, MA 02139 USA
  }

\begin{document}

\maketitle

\section{Abstract}
{
\bf

We explore the representational space of emotions by combining methods from different academic fields. Cognitive science has proposed appraisal theory as a view on human emotion with previous research showing how human-rated abstract event features can predict fine-grained emotions and capture the similarity space of neural patterns in mentalizing brain regions. At the same time, natural language processing (NLP) has demonstrated how transfer and multitask learning can be used to cope with scarcity of annotated data for text modeling.

The contribution of this work is to show that appraisal theory can be combined with NLP for mutual benefit. First, fine-grained emotion prediction can be improved to human-level performance by using NLP representations in addition to appraisal features. Second, using the appraisal features as auxiliary targets during training can improve predictions even when only text is available as input. Third, we obtain a representation with a similarity matrix that better correlates with the neural activity across regions. Best results are achieved when the model is trained to simultaneously predict appraisals, emotions and emojis using a shared representation. 

While these results are preliminary, the integration of cognitive neuroscience and NLP techniques opens up an interesting direction for future research.

}
\begin{quote}
\small
\textbf{Keywords:} 
appraisal theory; emotions; transfer learning 
\end{quote}

\vspace{-5 mm} 
\section{Introduction}

The representational space of emotions has been studied from several, generally disjoint approaches. On one hand, cognitive science has proposed principled and rich theories that involve fine-grained cognitive representations and causal dynamics \cite{scherer1999}. On the other, using datasets with thousands of examples and many parameters that normally don't involve semantically interpretable representations, machine learning has been making progress on natural language processing tasks like sentiment and emotion analysis. This work is a preliminary step in combining both approaches to improve emotion representation analysis from text. 

In situations limited by scarcity of data, humans rely on generalization from experience obtained in related tasks. Similarly in machine learning, parts of a model learned for one task can be transferred and reused in another related setting. Moreover, multi-tasking \cite{collobert}  learns useful representations for multiple purposes by jointly training a model on different simultaneous tasks. 


\begin{figure}[ht]
\begin{center}
\centerline{\includegraphics[scale=.35]{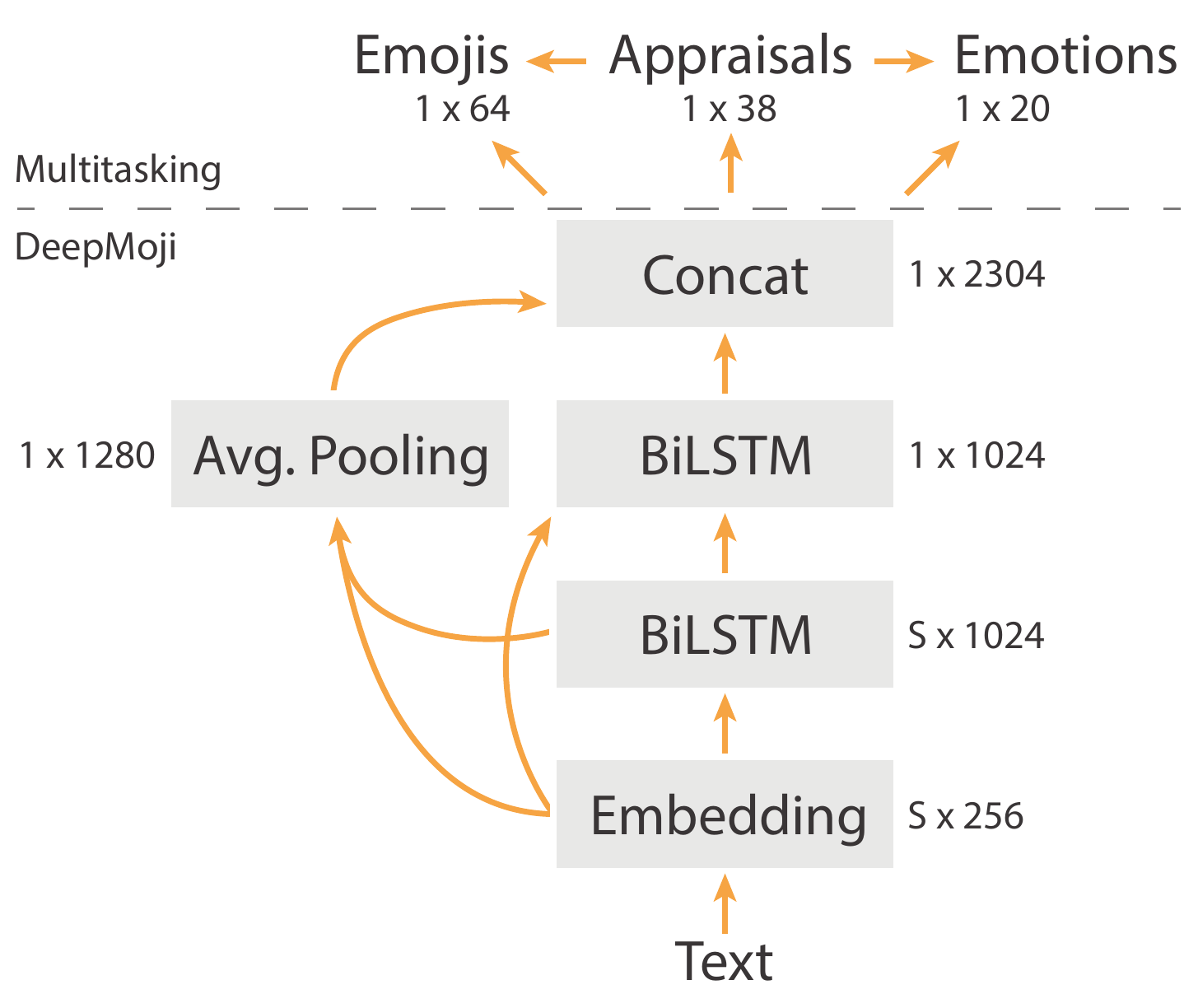}}
\end{center}
\vspace*{-8 mm}
\caption{Neural network architecture split into transferred layers of the existing model and our multitasking layers.} 
\label{architecture}
\end{figure}

\vspace{-2 mm} 
We build upon and combine the work of two previous papers. First, \citeA{skerry2015} identify a space of abstract situation features that well captures the fine-grained emotion discrimination that subjects make behaviorally on a set of 200 stimuli containing stories with emotional content. Moreover, they find that Theory of Mind (ToM) brain regions contains information about attributed emotions that can best be captured by the same space.  Second, Felbo et al. (2017) pretrain a deep learning model (DeepMoji) to predict emojis on a dataset with 634 million tweets containing 1 of 64 emojis, thereby learning a general representation of emotion that they use to obtain state-of-the-art performance on several benchmark datasets involving emotion, sarcasm and sentiment. 

\vspace{-2 mm} 
\section{Theory and Methods}

Among the alternative theories of the representational space of emotions, "appraisal theory" has been proposed as a way to capture the more intricate and fine-grained emotions that people experience and that people infer from what others experience. This theory characterizes emotions in terms of peoples interpretations or "appraisals" of the events around them. 

We use the stories and data from \citeA{skerry2015}. Subjects rated 200 stories with 38 appraisal features which of 20 emotion labels best described the emotion caused by each story (the subjects performed well above chance relative to the intended emotion, classifying the stimuli with 65\% accuracy). Multi-voxel pattern analysis was also performed on fMRI images from the ToM regions of the 22 subjects (in particular, we focus on the medial prefrontal cortex (MPFC) and on the right temporal parietal junction (RTPJ)).

To evaluate whether external NLP models could improve the emotional representation from \citeA{skerry2015}, we combined the appraisal features with features extracted from several natural language models. ``Word2Vec" is a bag-of-words model trained on the Google News Corpus~(Mikolov et al., 2013).``DeepMoji" was pretrained on Twitter data and its architecture consists of an embedding layer that learns a representation for each unique word, followed by two  bidirectional LSTM layers. Each LSTM layer has direct access to the embedding layer to allow more direct gradient flow. The representation obtained through the embedding layer is averaged and concatenated with the output of the LSTM layers as shown in Figure~\ref{architecture}\footnote{This architecture is from an early version of DeepMoji and differs slightly from the final architecture in the published paper.}. This ``Concat" layer contains the representation learned for the entire text and therefore contains the features used for the experiments in this paper.

\vspace{1 mm}
\subsubsection{MultiTask Learning}
We also tested a new representation by training on the simultaneous 3 tasks of reconstructing appraisals, predicting emojis and predicting emotions. To do this we fixed the parameters of DeepMoji and trained a subsequent layer of 38 dimensions (`appraisals') to reconstruct the abstract features with an mean square error loss. This layer, along with the `Concat' layer were used as input to two additional softmax layers trained to predict emojis and emotions, each with a categorical cross entropy loss. When training we alternated between predicting emojis using tweets and predicting emotions and appraisals using stories. 

\begin{table}[!th]
\vspace{-3 mm} 
\begin{center} 
\caption{Accuracy on emotion prediction. Reported mean accuracy when combining the extracted NLP representations and the appraisals at test time (`with appraisals') and when the appraisals are only available as targets during training for multitasking (`from text'). The 5\% confidence intervals are found across multiple balanced training-test splits\footnote{1000 times for all models except for multitasking, where we conducted the procedure 30 times due to computational constraints.}.} 
\label{sample-table} 
\vskip 0.12in
\begin{tabular}{L{1.8cm}C{2.3cm}C{2.3cm}} 
\hline
    &  With appraisals & From text \\
\hline
Chance & $.05$ & $.05$ \\
Word2Vec &   $.44  \pm .02$ &  $.10  \pm .01$\\
DeepMoji & $.53 \pm .05 $ & $.40  \pm .035$ \\
Multitasking &   $.64  \pm .04$ & $.42  \pm .05$  \\
\hline
Appraisals        &   $.58 \pm .003$ & --\\
Human-level           &   $.65$ & $.65$ \\
\hline
\end{tabular} 
\end{center} 
\label{experiment1}
\vspace*{-3 mm}
\end{table}

\vspace{-2 mm}
\section{Experiments and Results}

\subsubsection{Emotion Prediction} We tested the different models in their 20-way emotion classification capability both in combination with the manually annotated appraisal features and directly from text. A linear support vector machine (SVM) trained on appraisal features alone achieved 58\% while train on the combination of appraisal features and the multitasking representation achieved almost human level performance at 65\%. These results seem high given the small dataset size.

 \begin{figure}[!th]
\begin{center}
 \includegraphics[trim={2.5cm 0 5.2cm 0},clip, width=0.8\linewidth]{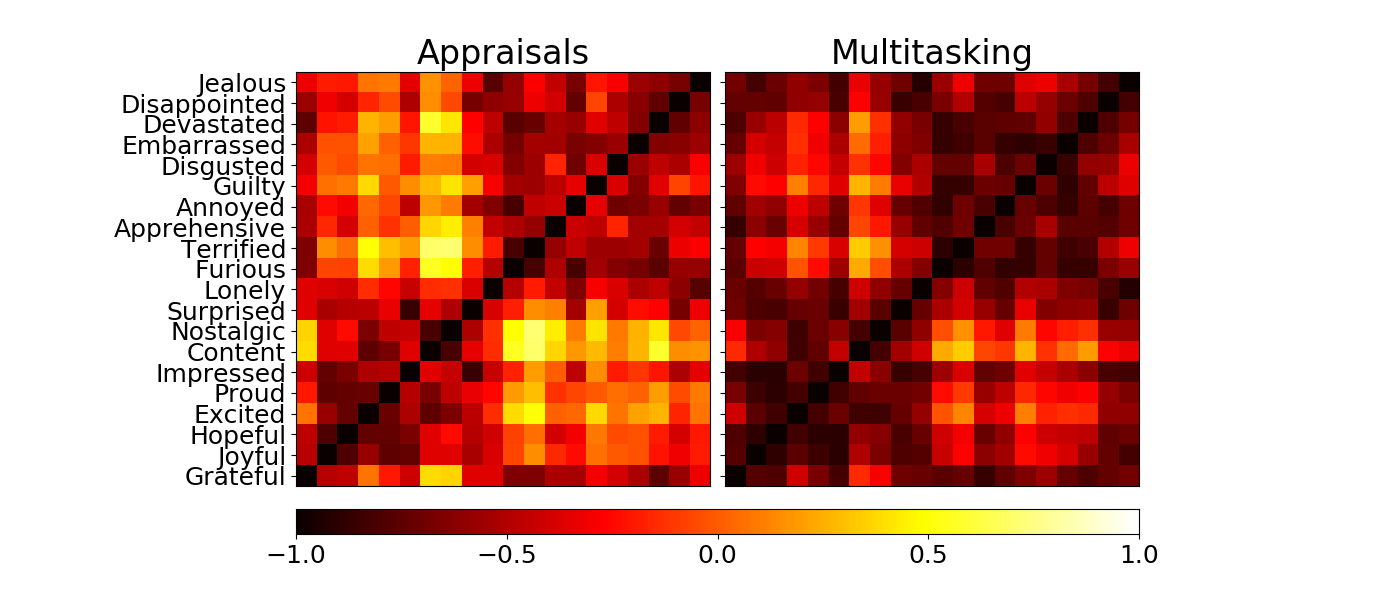}
\end{center}
\vspace*{-7 mm}
 \caption{Representational dissimilarity matrices (RDMs) encode the pairwise euclidean distance between different emotions within different feature spaces. The RDMs can be correlated to obtain a metric of similarity between  the spaces.} 
 \label{matrices}
 \vspace*{-3 mm}
 \end{figure}
 
\subsubsection{Representational Similarity Analysis}
We examine whether the improvement in classification of the representation learned in a multitasking way is reflected in a higher similarity with the structure of the representations on the ToM brain regions by comparing the representational dissimilarity matrices (RDMs) for the different neural and model feature spaces. We report results as the mean across multiple runs with different seeds. As shown in Figure~\ref{matrices}, the similarity of the representations learned through multitasking had a similar structure to that of the appraisal space (Kendall's tau of .74). The extent to which the feature spaces account for the similarity of the neural patterns is reported in Table 2. It does not escape our notice that the model trained for multitasking not only improves the classification of behavioral data, but also generates a closer match to the similarity space of the mentalizing brain regions than the human-labeled appraisal features alone.

 \begin{table}[!th]
 \vspace{-1 mm} 
\begin{center} 
\caption{Representational Similarity Analysis. Group-level Kendall's tau correlations between feature and neural RDMs.} 
\label{RDM} 
\vskip 0.12in
\begin{tabular}{lccc} 
\hline
    &  Appraisal & DeepMoji & Multitasking \\
\hline
DMPFC & .25 & .21 & .26  \\
MMPFC &  .19 &  .16 & .19\\
RTPJ & .24 &  .23 & .27\\
ToM &   .27 & .25 & .29 \\
\hline
\end{tabular} 
\end{center} 
\label{RDM}
\vspace*{-3 mm}
\end{table}

\vspace*{1 mm}

\nocite{bjarke2017}
\nocite{skerry2015}
\nocite{scherer1999}
\nocite{googleword2vec}

\bibliographystyle{apacite}

\setlength{\bibleftmargin}{.125in}
\setlength{\bibindent}{-\bibleftmargin}

\bibliography{ccn_style}
\end{document}